\documentclass[conference]{IEEEtran}
\IEEEoverridecommandlockouts

\usepackage{cite}
\usepackage{amsmath,amssymb,amsfonts}
\usepackage{graphicx}
\usepackage{textcomp}
\usepackage{xcolor}
\usepackage{booktabs}
\usepackage{balance}
\usepackage{tikz}
\usetikzlibrary{positioning,arrows.meta,fit,backgrounds,calc,shadows}
\usepackage{pgfplots}
\pgfplotsset{compat=1.18}
\usepackage{listings}
\definecolor{codegreen}{rgb}{0,0.6,0}
\definecolor{codegray}{rgb}{0.5,0.5,0.5}
\definecolor{codepurple}{rgb}{0.58,0,0.82}
\definecolor{backcolour}{rgb}{0.95,0.95,0.92}
\usepackage{stfloats}
\usepackage{capt-of}
\usepackage{hyperref}
\usepackage{xspace}
\newcommand{\eg}{e.g.,\xspace}

\def\BibTeX{{\rm B\kern-.05em{\sc i\kern-.025em b}\kern-.08em
    T\kern-.1667em\lower.7ex\hbox{E}\kern-.125emX}}

\newcommand{\numFwBenchmarks}{14}
\newcommand{\numFwModelServers}{six}
\newcommand{\numAuditCodebases}{six}
\newcommand{\numAuditBenchmarks}{three}
\newcommand{\numLbResults}{657}
\newcommand{\numLbBenchmarks}{17}
\newcommand{\numLbModels}{509}
\newcommand{\numLbPapersReviewed}{1{,}704}
\newcommand{\urlFramework}{https://github.com/allenai/vla-evaluation-harness}
\newcommand{\urlLeaderboard}{https://allenai.github.io/vla-evaluation-harness/leaderboard}

\begin{document}

\title{vla-eval: A Unified Evaluation Harness for\\Vision-Language-Action Models}

\author{
    \IEEEauthorblockN{Suhwan Choi\textsuperscript{1*},
        Yunsung Lee\textsuperscript{1},
        Yubeen Park\textsuperscript{1},
        Chris Dongjoo Kim\textsuperscript{2},
        Ranjay Krishna\textsuperscript{2},
        Dieter Fox\textsuperscript{2},
        Youngjae Yu\textsuperscript{3}}
    \IEEEauthorblockA{\textsuperscript{1}MAUM.AI \quad \textsuperscript{2}Allen Institute for AI (AI2) \quad \textsuperscript{3}Seoul National University\\
        \textsuperscript{*}Primary author}
}

\IEEEaftertitletext{%
    \begin{center}%
        \includegraphics[width=\textwidth]{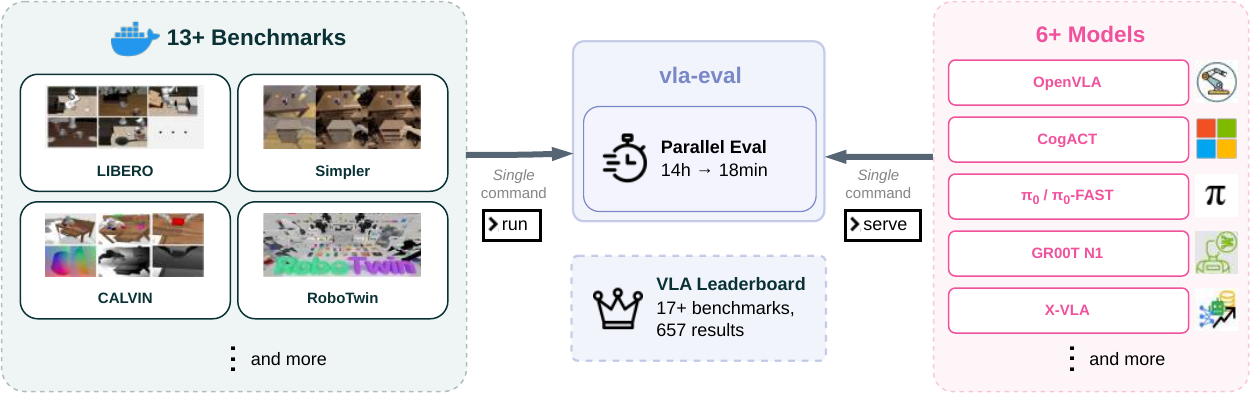}%
        \captionof{figure}{\label{fig:teaser}Overview of \texttt{vla-eval}: \numFwBenchmarks{} benchmarks (3 with full cross-codebase reproduction validation) and \numFwModelServers{} model servers each integrate once, requiring no per-benchmark dependency setup or manual asset installation, and connect through two commands (\texttt{run} and \texttt{serve}). The framework provides parallel evaluation (up to 47$\times$ speedup on LIBERO: 14h $\to$ 18min) and a VLA leaderboard aggregating \numLbResults{} results across \numLbBenchmarks{} benchmarks.}%
    \end{center}}

\maketitle

\begin{abstract}
    Vision-Language-Action (VLA) models are increasingly evaluated across multiple simulation benchmarks, yet adding each benchmark to an evaluation pipeline requires resolving incompatible dependencies, matching underspecified evaluation protocols, and reverse-engineering undocumented preprocessing. This burden scales with the number of models and benchmarks, making comprehensive evaluation impractical for most teams.
    We present \texttt{vla-eval}, an open-source evaluation harness that eliminates this per-benchmark cost by decoupling model inference from benchmark execution through a WebSocket+msgpack protocol with Docker-based environment isolation.
    Models integrate once by implementing a single \texttt{predict()} method; benchmarks integrate once via a four-method interface; the full cross-evaluation matrix works automatically.
    The framework supports \numFwBenchmarks{} simulation benchmarks and \numFwModelServers{} model servers.
    Parallel evaluation via episode sharding and batch inference achieves up to 47$\times$ wall-clock speedup, completing 2{,}000 LIBERO episodes in ${\sim}$18~minutes.
    To validate the framework, we reproduce published scores across \numAuditCodebases{} VLA codebases and \numAuditBenchmarks{} benchmarks, documenting previously undocumented pitfalls.
    We additionally release a VLA leaderboard aggregating \numLbResults{} published results across \numLbBenchmarks{} benchmarks.
    Framework, evaluation configs, and all reproduction results are publicly available.\footnote{\url{\urlFramework}}\textsuperscript{,}\footnote{\url{\urlLeaderboard}}
\end{abstract}


% ============================================================
\section{Introduction}

Recent Vision-Language-Action (VLA) models increasingly target multiple simulation benchmarks to demonstrate generalization across environments and embodiments~\cite{groot,pi05,xvla,dexbotic}.
However, adding even a single benchmark to an evaluation pipeline demands substantial engineering effort.
Each benchmark ships its own simulator, Python runtime, and asset requirements---LIBERO~\cite{libero} requires Python~3.8 with robosuite, ManiSkill2~\cite{maniskill2} requires Python~3.10 with SAPIEN, CALVIN~\cite{calvin} requires Python~3.8 with PyBullet---and no single environment can satisfy all constraints simultaneously.
Beyond dependency resolution, evaluation protocols are frequently underspecified: seeds, episode counts, and preprocessing details are omitted from papers, and a single undocumented parameter can shift success rates by up to 55 percentage points (Section~\ref{sec:audit}).
Correctly integrating one benchmark therefore requires not just environment setup but painstaking comparison against reference implementations.

This per-benchmark cost scales linearly: evaluating on $M$ benchmarks means repeating the process $M$ times, independently for each of $N$ models---an $O(N \times M)$ integration burden.
For small teams, comprehensive multi-benchmark evaluation is impractical.

We present \texttt{vla-eval} (Fig.~\ref{fig:teaser}), a unified evaluation harness that eliminates per-benchmark integration cost.
Following the decoupled design of \texttt{lm-evaluation-harness}~\cite{lmeval} for language models, \texttt{vla-eval} isolates each benchmark inside a Docker container and connects it to model servers via a WebSocket+msgpack protocol.
\textit{Models integrate once, benchmarks integrate once, and the full $N \times M$ cross-evaluation matrix works automatically}, reducing integration effort from $O(N \times M)$ to $O(N + M)$.
Our contributions are:
\begin{itemize}
    \item An open-source evaluation harness supporting \numFwBenchmarks{} benchmarks and \numFwModelServers{} model servers with Docker-based isolation and a WebSocket+msgpack protocol;
    \item Validation across \numAuditCodebases{} VLA codebases and \numAuditBenchmarks{} benchmarks, reproducing published scores and documenting pitfalls where a single undocumented parameter shifts success rates by up to 55 percentage points;
    \item A model-agnostic parallel evaluation methodology (episode sharding + batch inference) achieving up to 47$\times$ speedup, where the bottleneck is environment step rate rather than model inference;
    \item A VLA leaderboard with canonical protocol definitions, aggregating \numLbResults{} published results across \numLbBenchmarks{} benchmarks.
\end{itemize}

% ============================================================
\section{Framework Design}

\subsection{Architecture}

\texttt{vla-eval} separates model inference from benchmark execution via a client-server architecture using WebSocket with msgpack binary serialization.
Each message carries a type (\texttt{observation}, \texttt{action}, \texttt{episode\_start/end}), a benchmark-specific payload, a sequence number, and a timestamp.

\textbf{Model servers} extend \texttt{PredictModelServer}, which provides a blocking \texttt{predict(obs, ctx)} method (typically ${\sim}$50 lines), automatic action chunking, and optional batched inference via \texttt{max\_batch\_size}.
Listing~\ref{lst:openvla} shows the complete OpenVLA integration.

\begin{lstlisting}[language=Python, caption={OpenVLA model server (simplified).},label={lst:openvla},float=tb]
class OpenVLAServer(PredictModelServer):
  def __init__(self, model_path, **kw):
    super().__init__(**kw)
    self.model_path = model_path
    self._model = self._proc = None

  def _load_model(self):
    if self._model is not None:
      return
    self._proc = AutoProcessor.from_pretrained(
        self.model_path, trust_remote_code=True)
    self._model = AutoModelForVision2Seq \
        .from_pretrained(self.model_path,
          torch_dtype=torch.bfloat16,
          trust_remote_code=True).to("cuda")

  def predict(self, obs, ctx):
    self._load_model()
    img = Image.fromarray(
        next(iter(obs["images"].values())))
    prompt = f"In: What action should the robot" \
        f" take to {obs['task_description']}?\nOut:"
    inp = self._proc(prompt, img)
        .to("cuda", dtype=torch.bfloat16)
    act = self._model.predict_action(**inp)
    return {"actions": act}
\end{lstlisting}

\textbf{Dependency isolation.}
Each model server declares dependencies via PEP~723 inline metadata; \texttt{vla-eval serve} launches it through \texttt{uv run}, creating an isolated environment automatically.
Conflicting dependencies (\eg CogACT pinning \texttt{transformers==4.40.1} vs.\ X-VLA~\cite{xvla} requiring \texttt{transformers>=4.44}) coexist without interference, mirroring the Docker-based isolation used for benchmarks.

\textbf{Benchmarks} follow the same pattern: integrators implement four methods (\texttt{reset}, \texttt{step}, \texttt{make\_obs}, \texttt{get\_step\_result}) inside a dedicated Docker image with pinned dependencies.

\textbf{Declarative configs.}
Two YAML configs (benchmark + model server) drive each evaluation.
We publish all Docker images to \texttt{ghcr.io} with versioned tags and bundle all required assets (scene files, textures, robot descriptions), eliminating the ad-hoc asset installation that each benchmark otherwise requires.
A complete evaluation requires only two commands: \texttt{vla-eval serve} and \texttt{vla-eval run}.
% \begin{minipage}{\columnwidth}
%     \begin{lstlisting}[language={},caption={},label={},aboveskip=8pt]
% $ vla-eval serve --config model_server.yaml
% $ vla-eval run   --config benchmark.yaml
% \end{lstlisting}
% \end{minipage}
% TODO: Add a figure showing simplified benchmark + model server YAML config examples side by side.
Every run produces a structured JSON result file recording the harness version, benchmark configuration, and per-episode metrics, enabling exact reproduction.

\subsection{Supported Benchmarks and Models}

\begin{table}[tb]
    \centering
    \caption{Supported benchmarks. Docker = compressed image size; Act.\ = action space dimensionality; St.\ = status (C = cross-codebase reproduction verified, I = integrated but not yet cross-validated).}
    \label{tab:benchmarks}
    % Hidden metadata: Simulator / Robot
    % LIBERO: robosuite / Panda | CALVIN: PyBullet / Panda
    % SimplerEnv: SAPIEN / WidowX | ManiSkill2: SAPIEN / Panda
    % LIBERO-Mem: robosuite / Panda | Kinetix: Kinetix / ---
    % RoboCasa: robosuite / PandaOmron | VLABench: dm_control / Franka
    % MIKASA-Robo: ManiSkill3 / Panda | RoboTwin: SAPIEN / dual-arm
    % RLBench: CoppeliaSim / Panda | RoboCerebra: robosuite / Panda
    % LIBERO-Pro: robosuite / Panda
    \begin{tabular}{@{}lrcl@{}}
        \toprule
        Benchmark                      & Docker   & Act. & St. \\
        \midrule
        SimplerEnv~\cite{simplerenv}   & 4.9\,GB  & 7D   & C   \\
        LIBERO~\cite{libero}           & 6.0\,GB  & 7D   & C   \\
        CALVIN~\cite{calvin}           & 9.5\,GB  & 7D   & C   \\
        RLBench~\cite{rlbench}         & 4.7\,GB  & 8D   & I   \\
        LIBERO-Pro~\cite{liberopro}    & 6.2\,GB  & 7D   & I   \\
        RoboCerebra~\cite{robocerebra} & 6.3\,GB  & 7D   & I   \\
        ManiSkill2~\cite{maniskill2}   & 9.8\,GB  & 7D   & I   \\
        Kinetix~\cite{kinetix}         & 10.0\,GB & 6D   & I   \\
        MIKASA-Robo~\cite{mikasa}      & 10.1\,GB & 8D   & I   \\
        LIBERO-Mem~\cite{liberomem}    & 11.3\,GB & 7D   & I   \\
        RoboMME~\cite{robomme}         & 17.0\,GB & 8D   & I   \\
        VLABench~\cite{vlabench}       & 17.7\,GB & 7D   & I   \\
        RoboTwin~2.0~\cite{robotwin}   & 28.6\,GB & 14D  & I   \\
        RoboCasa~\cite{robocasa}       & 35.6\,GB & 7D   & I   \\
        \bottomrule
    \end{tabular}
\end{table}

Table~\ref{tab:benchmarks} lists all \numFwBenchmarks{} supported benchmarks with action spaces from 6D to 14D and Docker images from 4.7 to 35.6\,GB.
Model servers are implemented for \numFwModelServers{} models: CogACT~\cite{cogact}, OpenVLA~\cite{openvla}, OpenVLA-OFT~\cite{openvlaoft}, $\pi_0$~\cite{pi0}/$\pi_0$-FAST~\cite{pi0fast}, GR00T~N1~\cite{groot}, and X-VLA~\cite{xvla}.

\subsection{Parallel Evaluation}

\begin{figure}[tb]
    \centering
    \includegraphics[width=\columnwidth]{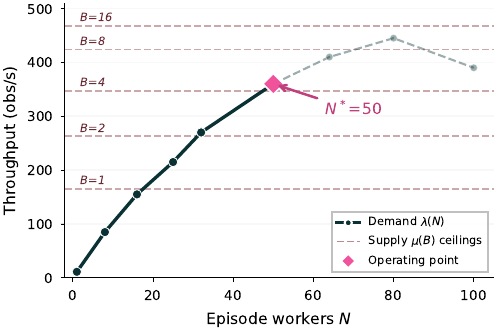}
    \caption{Demand/supply throughput for LIBERO + CogACT~\cite{cogact} on H100. Dashed lines show supply ceilings $\mu(B)$ at each batch size. The operating point $K^*\!=\!50$ uses 78\% of the supply capacity at $B\!=\!16$, leaving headroom to absorb burst arrivals and prevent queue buildup; beyond $K\!=\!80$, environment overhead causes throughput to drop.}
    \label{fig:throughput}
\end{figure}

Environment parallelism uses episode sharding across $K$ Docker containers; inference parallelism uses batched forward passes.
We tune parallelism via a demand/supply methodology (Fig.~\ref{fig:throughput}): $\lambda(K)$ measures environment throughput as a function of shards, $\mu(B)$ measures model throughput as a function of batch size, and the operating point satisfies $\lambda(K) < 0.8 \cdot \mu(B^*)$ to prevent queue buildup.

\textbf{Model-agnostic speedup.}
Model inference scales readily via batching, but existing benchmarks run a single environment instance, making simulation the dominant bottleneck.
Episode sharding closes this gap (Fig.~\ref{fig:throughput}): the model supply ceiling exceeds environment demand at all shard counts, so the speedup is determined by environment parallelism and transfers to any model.

\begin{figure}[tb]
    \centering
    \includegraphics[width=\columnwidth]{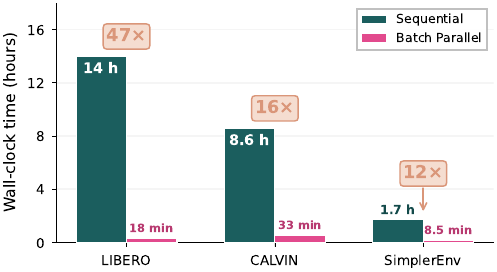}
    \caption{Wall-clock evaluation time: sequential vs.\ batch parallel. LIBERO: 2{,}000 episodes, 50 shards, $B\!=\!16$. CALVIN: 1{,}000 sequences, 16 shards. SimplerEnv: 288 episodes (3 seeds), 16 shards.}
    \label{fig:speedup}
\end{figure}

On LIBERO with CogACT-7B (H100 model server, separate benchmark host), episode sharding from $K\!=\!1$ to $K\!=\!50$ increases environment throughput by 32.6$\times$ ($\lambda$: 11.2$\to$364.6~observations per second (obs/s)), and batch inference from $B\!=\!1$ to $B\!=\!16$ increases model server throughput by 2.8$\times$ ($\mu$: 165.2$\to$468.2~obs/s).
Combined, 2{,}000 episodes complete in ${\sim}$18~minutes versus ${\sim}$14~hours sequentially, a \textbf{47$\times$ wall-clock speedup}.
The same methodology applies to CALVIN (1{,}000 sequences, 16~shards, ${\sim}$33~min, 16$\times$ speedup) and SimplerEnv (288 episodes, 16~shards, ${\sim}$8.5~min, 12$\times$ speedup), as shown in Fig.~\ref{fig:speedup}.

% ============================================================
\section{Validation}
\label{sec:audit}

\subsection{Scope and Results}

To validate the framework, we evaluate \numAuditCodebases{} published VLA codebases---OpenVLA~\cite{openvla}, $\pi_{0.5}$~\cite{pi05}, OpenVLA-OFT~\cite{openvlaoft}, GR00T~N1.6~\cite{groot}, DB-CogACT~\cite{dexbotic}, and X-VLA~\cite{xvla}---across three simulation benchmarks using fixed seeds and versioned Docker images from \texttt{ghcr.io}.
Most VLA models require per-benchmark fine-tuning, so evaluation is only possible where public checkpoints exist.
LIBERO: 4 suites $\times$ 10 tasks $\times$ 50 episodes (2{,}000 total).
CALVIN: ABC$\to$D, 1{,}000 chained sequences.
SimplerEnv: 4 WidowX tasks, 24 episodes per task.

\begin{table}[tb]
    \centering
    \caption{Reproduction matrix: ours ($\Delta$ vs.\ reported).}
    \label{tab:repro}
    \begin{tabular}{@{}lccc@{}}
        \toprule
        Codebase    & LIBERO (\%)                        & CALVIN (len)                 & SimplerEnv (\%)                     \\
        \midrule
        OpenVLA     & 76.2\,{\scriptsize($-$0.3)}        & ---                          & ---                                 \\
        $\pi_{0.5}$ & 97.7\,{\scriptsize(+0.8)}          & ---                          & ---                                 \\
        OpenVLA-OFT & 96.7\,{\scriptsize($-$0.4)}        & ---                          & ---                                 \\
        GR00T~N1.6  & 94.9\,{\scriptsize($-$2.1)}$^\dag$ & ---                          & 59.7\,{\scriptsize($-$8.0)}$^\ddag$ \\
        DB-CogACT   & 94.7\,{\scriptsize($-$0.2)}        & 4.02\,{\scriptsize($-$0.04)} & 63.5\,{\scriptsize($-$6.0)}         \\
        X-VLA       & 97.4\,{\scriptsize($-$0.7)}        & 4.30\,{\scriptsize($-$0.13)} & 94.8\,{\scriptsize($-$1.0)}         \\
        \bottomrule
    \end{tabular}\\[2pt]
    {\scriptsize --- = no public checkpoint.
    $^\dag$Community checkpoint.
    $^\ddag$Google Robot visual matching (others are WidowX).}
\end{table}

Table~\ref{tab:repro} shows the reproduction matrix.
Published scores largely reproduce across \numAuditCodebases{} codebases and \numAuditBenchmarks{} benchmarks, validating the framework's fidelity to reference implementations.

\subsection{Reproduction Challenges}

These reproductions were non-trivial: single undocumented settings could cause catastrophic score changes.

Using the wrong proprioceptive state source in X-VLA~\cite{xvla} on LIBERO drops success rate from 97.8\% to 42\%, a 55 percentage point (pp) swing from one parameter.
Confusing absolute and delta action modes (both valid 7D vectors, indistinguishable from data alone) produces 0\% as positions accumulate and the robot diverges.
OpenVLA-OFT~\cite{openvlaoft} uses a quaternion-to-axis-angle conversion without antipodal normalization (angle $\in [0, 2\pi]$, matching robosuite convention), while our initial implementation flipped $w<0$ quaternions (angle $\in [0, \pi]$); this single mismatch dropped LIBERO-Goal from 97\% to 83\% and LIBERO-Long from 95\% to 56\%.
OpenVLA~\cite{openvla} applies a center crop (scale\,=\,0.9) at evaluation time that is not documented in the paper; omitting it costs ${\sim}$3\,pp.
GR00T~\cite{groot} expects end-effector pose as proprioceptive input, but this field exists only in an internal simulator fork, not in official SimplerEnv~\cite{simplerenv}; without it, scores drop from 30--55\% to 0\%.

Each of these was discovered only through systematic comparison of intermediate values against reference implementations.
Reimplementing the necessary patches brought GR00T on SimplerEnv (Google Robot) from 0\% to 59.7\%, with a $-$8.0\,pp gap to the reported score remaining.

% ============================================================
\section{VLA Leaderboard}

\begin{figure*}[t]
    \centering
    \includegraphics[width=0.94\textwidth]{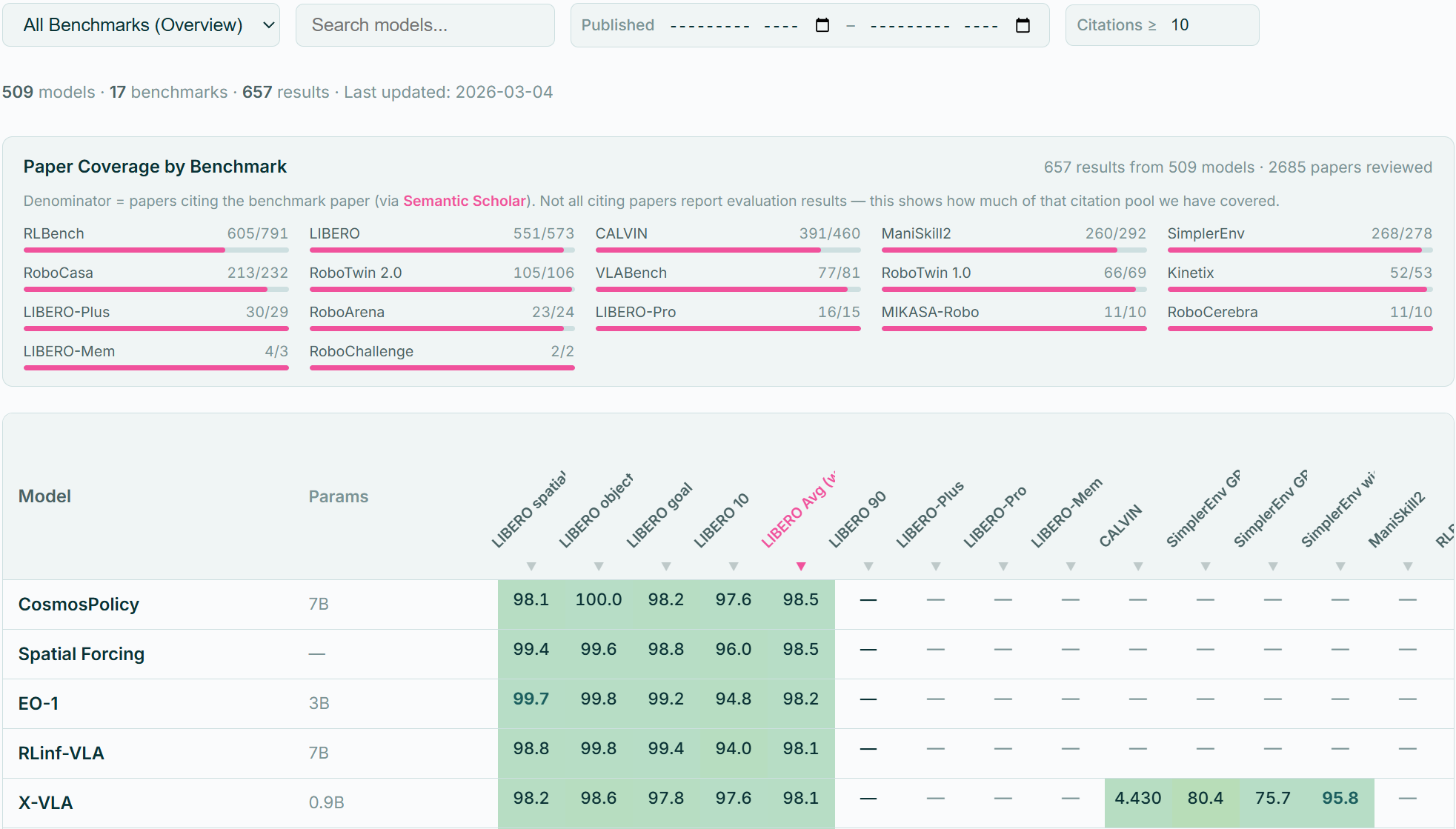}
    \caption{VLA leaderboard (\numLbBenchmarks{} benchmarks, \url{\urlLeaderboard}). Shown: models with $>$10 citations. Filterable by benchmark and model.}
    \label{fig:leaderboard}
\end{figure*}

Beyond framework validation, we compile a broader picture of the VLA evaluation landscape.
We release a VLA leaderboard (Fig.~\ref{fig:leaderboard}) aggregating \numLbResults{} results across \numLbBenchmarks{} benchmarks and \numLbModels{}+ configurations, sourced from \numLbPapersReviewed{} papers that cite at least one of the tracked benchmarks.

\subsection{Curation}

Evaluation protocols vary across papers: SimplerEnv spans three incomparable robot configurations; CALVIN ABC$\to$D and ABCD$\to$D splits are not comparable; LIBERO papers report 4 or 5 suites.
We established canonical protocol definitions for each benchmark, standardizing task subsets, metrics, splits, and comparability constraints.

An AI agent (Claude Code with Opus~4.6) reviewed \numLbPapersReviewed{} papers via MCP tool integrations (arXiv, Semantic Scholar, PDF reader) to extract and normalize results against these canonical protocols.
A human operator then reviewed every entry, resolving anomalies and ambiguous cases.
Each entry is versioned with full provenance metadata and validated against automated schema constraints, enabling fair cross-paper comparison.

\subsection{Cross-Benchmark Analysis}

\begin{figure}[tb]
    \centering
    \includegraphics[width=\columnwidth]{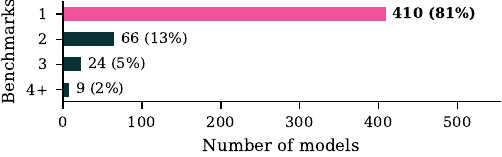}
    \caption{Distribution of benchmark coverage per model. 81\% of the \numLbModels{}+ models are evaluated on only one benchmark; only 3 (0.6\%) on 5 or more.}
    \label{fig:eval-coverage}
\end{figure}

Fig.~\ref{fig:eval-coverage} shows the distribution of benchmark coverage across \numLbModels{}+ models and the \numLbBenchmarks{} benchmarks tracked in the leaderboard: 81\% are evaluated on a single benchmark, and only 6\% on three or more.
Cross-benchmark comparison is therefore rare, limiting our ability to assess general model capability across diverse environments and embodiments.
This underscores the need for a unified framework that makes cross-benchmark comparison practical.

% ============================================================
\section{Conclusion}

Validation across \numAuditCodebases{} codebases and \numAuditBenchmarks{} benchmarks confirms that \texttt{vla-eval} reproduces published scores within expected variance, while revealing that single undocumented parameters can shift results by tens of pp.
\texttt{vla-eval} records the full evaluation configuration alongside every result, making any run reproducible from a single config file.

\textbf{Limitations.}
Our audit covers \numAuditCodebases{} codebases across \numAuditBenchmarks{} simulation benchmarks; additional benchmarks and real-robot transfer are planned.
Leaderboard results are extracted from published papers, not independently verified.
Supported metrics are limited to task success rate; finer-grained dimensions such as subtask progress, task efficiency, and safety are not yet supported.

\balance

\section*{Acknowledgments}

This work was partly supported by an Institute of Information \& Communications Technology Planning \& Evaluation (IITP) grant funded by the Korean Government (MSIT) (No. RS-2021-II211343, Artificial Intelligence Graduate School Program (Seoul National University)), and by the Technology Innovation Program (RS-2025-25456760, Development of a Humanoid Robot Specialized in Chemical Processes Based on an AI Foundation Model), funded by the Ministry of Trade, Industry and Resources (MOTIR), Korea.

\bibliographystyle{IEEEtran}
\bibliography{references}

@article{dexbotic,
  title   = {{Dexbotic}: Open-Source Vision-Language-Action Toolbox},
  author  = {Xie, Bin and Zhou, Erjin and Jia, Fan and Shi, Hao and Fan, Haoqiang and Zhang, Haowei and others},
  journal = {arXiv preprint arXiv:2510.23511},
  year    = {2025}
}

@inproceedings{openvla,
  title     = {{OpenVLA}: An Open-Source Vision-Language-Action Model},
  author    = {Kim, Moo Jin and Pertsch, Karl and Karamcheti, Siddharth and Xiao, Ted and Balakrishna, Ashwin and Nair, Suraj and others},
  booktitle = {CoRL},
  year      = {2024}
}

@article{pi0,
  title   = {$\pi_0$: A Vision-Language-Action Flow Model for General Robot Control},
  author  = {Black, Kevin and Brown, Noah and Driess, Danny and Esmail, Adnan and Equi, Michael and Finn, Chelsea and others},
  journal = {arXiv preprint arXiv:2410.24164},
  year    = {2024}
}

@article{cogact,
  title   = {{CogACT}: A Foundational Vision-Language-Action Model for Synergizing Cognition and Action in Robotic Manipulation},
  author  = {Li, Qixiu and Liang, Yaobo and Wang, Zeyu and Luo, Lin and Chen, Xi and Liao, Mozheng and others},
  journal = {arXiv preprint arXiv:2411.19650},
  year    = {2024}
}

@article{groot,
  title   = {{GR00T} {N1}: An Open Foundation Model for Generalist Humanoid Robots},
  author  = {Bjorck, Johan and Casta{\~n}eda, Fernando and Cherniadev, Nikita and Da, Xingye and Ding, Runyu and Fan, Linxi and others},
  journal = {arXiv preprint arXiv:2503.14734},
  year    = {2025}
}

@article{pi0fast,
  title   = {{FAST}: Efficient Action Tokenization for Vision-Language-Action Models},
  author  = {Pertsch, Karl and Stachowicz, Kyle and Ichter, Brian and Driess, Danny and Nair, Suraj and Vuong, Quan and others},
  journal = {arXiv preprint arXiv:2501.09747},
  year    = {2025}
}

@article{openvlaoft,
  title   = {Fine-Tuning Vision-Language-Action Models: Optimizing Speed and Success},
  author  = {Kim, Moo Jin and Finn, Chelsea and Liang, Percy},
  journal = {arXiv preprint arXiv:2502.19645},
  year    = {2025}
}

@article{xvla,
  title   = {{X-VLA}: Soft-Prompted Transformer as Scalable Cross-Embodiment Vision-Language-Action Model},
  author  = {Zheng, Jinliang and Li, Jianxiong and Wang, Zhihao and Liu, Dongxiu and Kang, Xirui and Feng, Yuchun and others},
  journal = {arXiv preprint arXiv:2510.10274},
  year    = {2025}
}

@article{robomme,
  title   = {{RoboMME}: Benchmarking and Understanding Memory for Robotic Generalist Policies},
  author  = {Dai, Yinpei and Fu, Hongze and Lee, Jayjun and Liu, Yuejiang and Zhang, Haoran and Yang, Jianing and others},
  journal = {arXiv preprint arXiv:2603.04639},
  year    = {2026}
}

@inproceedings{libero,
  title     = {{LIBERO}: Benchmarking Knowledge Transfer for Lifelong Robot Learning},
  author    = {Liu, Bo and Zhu, Yifeng and Gao, Chongkai and Feng, Yihao and Liu, Qiang and Zhu, Yuke and others},
  booktitle = {NeurIPS Datasets and Benchmarks},
  year      = {2023}
}

@article{calvin,
  title   = {{CALVIN}: A Benchmark for Language-Conditioned Policy Learning for Long-Horizon Robot Manipulation Tasks},
  author  = {Mees, Oier and Hermann, Luk\'{a}s and Rosete-Beas, Erick and Burgard, Wolfram},
  journal = {IEEE Robotics and Automation Letters},
  year    = {2022}
}

@inproceedings{simplerenv,
  title     = {Evaluating Real-World Robot Manipulation Policies in Simulation},
  author    = {Li, Xuanlin and Hsu, Kyle and Gu, Jiayuan and Mees, Oier and Pertsch, Karl and Walke, Homer Rich and others},
  booktitle = {CoRL},
  year      = {2024}
}

@inproceedings{maniskill2,
  title     = {{ManiSkill2}: A Unified Benchmark for Generalizable Manipulation Skills},
  author    = {Gu, Jiayuan and Xiang, Fanbo and Li, Xuanlin and Ling, Zhan and Liu, Xiqiang and Mu, Tongzhou and others},
  booktitle = {ICLR},
  year      = {2023}
}

@article{liberomem,
  title   = {Rethinking Progression of Memory State in Robotic Manipulation: An Object-Centric Perspective},
  author  = {Chung, Nhat and Hanyu, Taisei and Nguyen, Toan and Le, Huy and Bumgarner, Frederick and Nguyen, Duy Minh Ho and others},
  journal = {arXiv preprint arXiv:2511.11478},
  year    = {2025}
}

@inproceedings{kinetix,
  title     = {{Kinetix}: Investigating the Training of General Agents through Open-Ended Physics-Based Control Tasks},
  author    = {Matthews, Michael and Beukman, Michael and Lu, Chris and Foerster, Jakob},
  booktitle = {ICLR},
  year      = {2025}
}

@inproceedings{robocasa,
  title     = {{RoboCasa}: Large-Scale Simulation of Household Tasks for Generalist Robots},
  author    = {Nasiriany, Soroush and Maddukuri, Abhiram and Zhang, Lance and Parikh, Adeet and Lo, Aaron and Joshi, Abhishek and others},
  booktitle = {RSS},
  year      = {2024}
}

@article{vlabench,
  title   = {{VLABench}: A Large-Scale Benchmark for Language-Conditioned Robotics Manipulation with Long-Horizon Reasoning Tasks},
  author  = {Zhang, Shiduo and Xu, Zhe and Liu, Peiju and Yu, Xiaopeng and Li, Yuan and Gao, Qinghui and others},
  journal = {arXiv preprint arXiv:2412.18194},
  year    = {2024}
}

@article{mikasa,
  title   = {Memory, Benchmark \& Robots: A Benchmark for Solving Complex Tasks with Reinforcement Learning},
  author  = {Cherepanov, Egor and Kachaev, Nikita and Kovalev, Alexey K. and Panov, Aleksandr I.},
  journal = {arXiv preprint arXiv:2502.10550},
  year    = {2025}
}

@article{robotwin,
  title   = {{RoboTwin} 2.0: A Scalable Data Generator and Benchmark with Strong Domain Randomization for Robust Bimanual Robotic Manipulation},
  author  = {Chen, Tianxing and Chen, Zanxin and Chen, Baijun and Cai, Zijian and Liu, Yibin and Li, Zixuan and others},
  journal = {arXiv preprint arXiv:2506.18088},
  year    = {2025}
}

@article{rlbench,
  title   = {{RLBench}: The Robot Learning Benchmark \& Learning Environment},
  author  = {James, Stephen and Ma, Zicong and Arrojo, David Rovick and Davison, Andrew J.},
  journal = {IEEE Robotics and Automation Letters},
  year    = {2020}
}

@inproceedings{robocerebra,
  title     = {{RoboCerebra}: A Large-scale Benchmark for Long-horizon Robotic Manipulation Evaluation},
  author    = {Han, Songhao and Qiu, Boxiang and Liao, Yue and Huang, Siyuan and Gao, Chen and Yan, Shuicheng and others},
  booktitle = {NeurIPS Datasets and Benchmarks},
  year      = {2025}
}

@article{liberopro,
  title   = {{LIBERO-PRO}: Towards Robust and Fair Evaluation of Vision-Language-Action Models Beyond Memorization},
  author  = {Zhou, Xueyang and Xu, Yangming and Tie, Guiyao and Chen, Yongchao and Zhang, Guowen and Chu, Duanfeng and others},
  journal = {arXiv preprint arXiv:2510.03827},
  year    = {2025}
}

@article{pi05,
  title   = {$\pi_{0.5}$: a Vision-Language-Action Model with Open-World Generalization},
  author  = {Black, Kevin and Brown, Noah and Darpinian, James and Dhabalia, Karan and Driess, Danny and Esmail, Adnan and others},
  journal = {arXiv preprint arXiv:2504.16054},
  year    = {2025}
}

@misc{lmeval,
  title     = {The Language Model Evaluation Harness},
  author    = {Gao, Leo and Tow, Jonathan and Abbasi, Baber and Biderman, Stella and Black, Sid and DiPofi, Anthony and others},
  year      = {2024},
  doi       = {10.5281/zenodo.12608602},
  publisher = {Zenodo}
}

\end{document}